\documentclass[nohyperref]{article}

\usepackage{microtype}
\usepackage{graphicx}
\usepackage{subfigure}
\usepackage{booktabs} 

\usepackage{hyperref}

\usepackage[accepted]{icml2022}

\usepackage{amsmath}
\usepackage{amssymb}
\usepackage{mathtools}
\usepackage{amsthm}
\usepackage{multirow}
\usepackage{appendix}

\usepackage[capitalize,noabbrev]{cleveref}

\theoremstyle{plain}
\newtheorem{theorem}{Theorem}[section]
\newtheorem{proposition}[theorem]{Proposition}
\newtheorem{lemma}[theorem]{Lemma}

\theoremstyle{definition}
\newtheorem{definition}[theorem]{Definition}

\theoremstyle{remark}
\newtheorem{remark}[theorem]{Remark}

\usepackage[textsize=tiny]{todonotes}

\icmltitlerunning{Rethinking Multidimensional Discriminator Output}

\begin{document}

\twocolumn[
\icmltitle{Rethinking Multidimensional Discriminator Output \\
for Generative Adversarial Networks}



\icmlsetsymbol{equal}{*}

\begin{icmlauthorlist}
\icmlauthor{Mengyu Dai}{equal,comp}
\icmlauthor{Haibin Hang}{equal,yyy}
\icmlauthor{Anuj Srivastava}{sch}
\end{icmlauthorlist}

\icmlaffiliation{comp}{Salesforce}
\icmlaffiliation{yyy}{Amazon}
\icmlaffiliation{sch}{Department of Statistics, Florida State University}

\icmlcorrespondingauthor{Mengyu Dai}{mdai@salesforce.com}
\icmlcorrespondingauthor{Haibin Hang}{haibinh@amazon.com}
\icmlcorrespondingauthor{Anuj Srivastava}{anuj@stat.fsu.edu}


\vskip 0.3in
]



\printAffiliationsAndNotice{\icmlEqualContribution} 

\begin{abstract}
The study of multidimensional discriminator (critic) output for Generative Adversarial Networks has been underexplored in the literature.
In this paper, we generalize the Wasserstein GAN framework to take advantage of multidimensional critic output  
and explore its properties.
We also introduce a square-root velocity transformation (SRVT) block which favors training in the multidimensional setting.
Proofs of properties are based on our proposed maximal $p$-centrality discrepancy, which is bounded above by $p$-Wasserstein distance and fits the Wasserstein GAN framework with multidimensional critic output $n$. Especially when $n=1$ and $p=1$, the proposed discrepancy equals $1$-Wasserstein distance.
Theoretical analysis and empirical evidence show that high-dimensional critic output has its advantage on distinguishing real and fake distributions, and benefits faster convergence and diversity of results.
\end{abstract}

\section{Introduction}\label{sec:introduction}
Generative Adversarial Networks (GAN) have led to numerous success stories in various tasks in recent years~\cite{yang2022surrealgansemisupervised,yu2022generating,Niemeyer_2021_CVPR, Chan_2021_CVPR, Han_Chen_Liu_2021, Karras2020ada, nauata2020house, Heim_2019_CVPR}. 
The goal in a GAN framework is to learn a distribution (and generate fake data) that is as close to real data distribution as possible. This is achieved by playing a two-player game, in which a generator and a discriminator compete with each other and try to reach a Nash equilibrium~\cite{goodfellow2014generative}. Arjovsky et al. ~\cite{arjovsky2017principled,wgan2017} pointed out the shortcomings of using Jensen-Shannon Divergence in formulating the objective function, and proposed using the $1$-Wasserstein distance instead. Numerous promising frameworks \cite{li2017mmd, mcgan2017,mroueh2017fisher,sobolevgan2017,Wu_2019_CVPR,Deshpande_2019_CVPR,Ansari_2020_CVPR} based on other discrepancies were developed afterwards. 
Although some of these works use critic output dimension $n=1$, empirical evidence can be found that using multiple dimension $n$ could be advantageous. 
For examples, in \cite{li2017mmd} authors pick different $n$s ($16,64,128$) for different datasets; In Sphere GAN \cite{Park_2019_CVPR} their ablation study shows the best performance with $n=1024$. However, the reason for this phenomenon has not been well explored yet.

One contribution of this paper is to explore the properties of multidimensional critic output in the generalized WGAN framework. Particularly, we propose a new metric on the space of probability distributions, called {\it maximal $p$-centrality discrepancy}. This metric is closely related to $p$-Wasserstein distance (Theorem~\ref{T:main}) and can serve as an alternative of WGAN objective especially when the discriminator has multidimensional output. 
In this revised WGAN framework we show that using high-dimensional critic output could make discriminator more informative on distinguishing real and fake distributions (Proposition~\ref{P:high-dim}). In classical WGAN with only one critic output, the discriminator push-forwards (or projects) real and fake distributions to $1$-dimensional space, and then look at their maximal mean discrepancy. 
This $1$-dimensional push-forward may hide significant differences of distributions in the shadow. Even though ideally there exists a ``perfect'' push-forward which reveals any tiny differences, practically the discriminator has difficulties to reach that global optimal push-forward~\cite{Stanczuk2021WassersteinGW}. 
However, using $p$-centrality allows to push-forward distributions to higher dimensional space. Since even an average high-dimensional push-forward may reveal more differences than a good $1$-dimensional push-forward, this reduces the burden on discriminator. Specifically, we show that more faithful $p$-centrality functions returns larger discrepancies between probability distributions (Lemma~\ref{P:high-dim}).

Another novelty of this work is to break the symmetry structure of the discriminator network by compositing with an asymmetrical square-root velocity transformation (SRVT). In general architectures people assume that the output layer of discriminator is fully connected. This setup puts all output neurons in equal and symmetric positions. As a result, any permutation of the multidimensional output vector will leave the value of objective function unchanged. This permutation symmetry implies that the weights connected to output layer are somehow correlated 
and this would undermine the generalization power of the discriminator network \cite{liang2019fisher,badrinarayanan2015understanding}. 
After adding the asymmetrical SRVT block, each output neuron would be structurally unique (Proposition~\ref{P:unique}).
Our understanding is that the structural uniqueness of output neurons would imply their functionality uniqueness. This way, different output neurons are forced to reflect distinct features of input distribution. Hence SRVT serves as an magnifier which favors the use of high-dimensional critic output.
\section{Related Work}
\noindent
{\bf Wasserstein Distance and Other Discrepancies Used in GAN:}
Arjovsky et al. \cite{wgan2017} applied Kantorovich-Rubinstein duality for $1$-Wasserstein distance as loss function in GAN objective. 
WGAN makes great progress toward stable training compared with previous GANs, and marks the start of using Wasserstein distance in GAN. However, sometimes it still may converge to sub-optimal optima or fail to converge due to the raw realization of Lipschitz condition by weight clipping. 
To resolve these issues, researchers proposed sophisticated ways\cite{gulrajani2017improved, wei2018improving, miyato2018spectral} to enforce Lipschitz condition for stable training.
Recently, people come up with another way to involve Wasserstein distance in GAN~\cite{Wu_2019_CVPR,NEURIPS2019_f0935e4c,Deshpande_2018_CVPR,Lee_2019_CVPR}. They use the  Sliced Wasserstein Distance~\cite{rabin2011wasserstein,kolouri2016sliced} to estimate the Wasserstein distance from samples based on a summation over the projections along random directions. Either of these methods rely on pushforwards of real and fake distributions through Lipschitz functions or projections on to $1$-dimensional space. In our work, we attempt to distinguish two distributions by looking at their pushforwards in high dimensional space. 



Another way people used to distinguish real data and fake data distributions in generative network is by moment matching~\cite{gmmn2015,dziugaite2015training}. Particularly, in~\cite{li2017mmd} the authors used the kernel maximum mean discrepancy (MMD) in GAN objective, which aims to match infinite order of moments. 
In our work we propose to use the maximum discrepancy between $p$-centrality functions to measure the distance of two distributions. The $p$-centrality function (Definition~\ref{D:p-centrality}) is exactly the $p$-th root of the $p$-th moment of a distribution. Hence, the maximal $p$-centrality discrepancy distance we propose can be viewed as an attempt to match the $p$-th moment for any given $p\geq 1$.

\noindent
{\bf $p$-Centrality Functions:}
The mean or expectation of a distribution is a basic statistic. Particularly, in Euclidean spaces, it is well known that the mean realizes the unique minimizer of the so-called Fr\'{e}chet function of order $2$ (cf.\,\cite{grove1973conjugatec,bhattacharya2003large,arnaudon2013medians}). Generally speaking, a Fr\'{e}chet function of order $p$ summarizes the $p$-th moment of a distribution with respect to any base point. A topological study of Fr\'{e}chet functions is carried out in~\cite{hang2019topological} which shows that by taking $p$-th root of a Fr\'{e}chet function, the $p$-centrality function can derive topological summaries of a distribution which is robust with respect to $p$-Wasserstein distance. In our work, we propose using $p$-centrality functions to build a nice discrepancy distance between distributions, which would benefit from its close connection with $p$-Wasserstein distance.

\noindent
{\bf Asymmetrical Networks:} 
Symmetries occur frequently in deep neural networks. By symmetry we refer to certain group actions on the weight parameter space which keep the objective function invariant. These symmetries would cause redundancy in the weight space and affects the generalization capacity of network \cite{liang2019fisher,badrinarayanan2015understanding}.
There are two types of symmetry: (i) permutation invariant; (ii) rescaling invariant.
A straight forward way to break symmetry is by random initialization (cf.~\cite{glorot2010understanding,he2015delving}).
Another way to break symmetry is via skip connections to add extra connections between nodes in different layers~\cite{he2016deep,he2016identity,huang2017densely}.
In our work, we attempt to break the permutation symmetry of the output layer in the discriminator using a nonparametric asymmetrical transformation specified by square-root velocity function (SRVF) \cite{5601739, srivastava2016functional}. 
The simple transformation that converts functions into their SRVFs changes Fisher-Rao metric into the $L^2$ norm, enabling efficient analysis of high-dimensional data.
Since the discretised formulation of SRVF is equivalent with an non-fully connected network, 
it can be viewed as breaking symmetry by deleting specific connections from the network.

\section{Methodology}
In this section we use the proposed GAN framework as a starting point 
to study the behaviors of multidimensional critic output. Proofs of concepts can be found in Appendix.
\subsection{Objective Function}
The objective function of the proposed GAN is as follows:
\begin{equation}
\label{eqn:obj}
\min\limits_{G} \max\limits_{D} \big(E_x [\|D(x)\|^p]\big)^{1/p} - \big(E_z [\|D(G(z))\|^p]\big)^{1/p}
\end{equation}
where $\| \cdot \|$ denotes $L^2$ norm. $G$ and $D$ denotes generator and discriminator respectively. $p$ refers to the order of moments.  $x \sim \mathbb{P}_r$ is the input real sample and $z \sim p(z)$ is a noise vector for the generated sample. The output of the last dense layer of discriminator is an $n$-dimensional vector in the Euclidean space $\mathbb{R}^n$. 
In contrast to traditional WGAN with $1$-dimensional discriminator output, our framework allows the last dense layer of discriminator to have multidimensional output. 

\subsection{$p$-centrality function}
The $p$-centrality function was introduced in~\cite{hang2019topological} which offers a way to obtain robust topological summaries of a probability distribution. 
In this section we show that $p$-centrality function is not only a robust but also a relatively faithful indicator of a probability distribution. 
\begin{definition}[$p$-centrality function]\label{D:p-centrality}
Given a Borel probability measure $\mathbb{P}$ on a metric space $(M,d)$ and $p\geq 1$, the $p$-centrality function is defined as
\begin{align*}
    \sigma_{\mathbb{P},p}(x) :=& \left(\int_M d^p(x,y)d\mathbb{P}(y)\right)^{\frac{1}{p}}
    =\left(\mathbb{E}_{y\sim\mathbb{P}}[d^p(x,y)]\right)^{\frac{1}{p}}.
\end{align*}
\end{definition}
Particularly, the value of $p$-centrality function at $x$ is the $p$-th root of the $p$-th moment of $\mathbb{P}$ with respect to $x$. As we know it, the $p$-th moments are important statistics of a probability distribution. After taking the $p$-th root, the $p$-centrality function retains those important information in $p$-th moments, and it also shows direct connection with the $p$-Wasserstein distance $W_p$:
\begin{lemma}\label{L:dirac}
For any $x\in M$, let $\delta_x$ be the Dirac measure centered at $x$. Then $\sigma_{\mathbb{P},p}(x)=W_p(\mathbb{P},\delta_x)$.
\end{lemma}

\begin{lemma}\label{L:triangle}
For any two Borel probability measures $\mathbb{P}$ and $\mathbb{Q}$ on $(M,d)$, we have
\begin{align*}
    \|\sigma_{\mathbb{P},p}-\sigma_{\mathbb{Q},p}\|_\infty\leq W_p(\mathbb{P},\mathbb{Q})\leq \|\sigma_{\mathbb{P},p}+\sigma_{\mathbb{Q},p}\|_\infty.
\end{align*}
\end{lemma}
Let $\mathcal{P}(M)$ be the set of all probability measures on $M$ and let $C_0(M)$ be the set of all continuous functions on $M$. We define an operator $\Sigma_p:\mathcal{P}(M)\rightarrow C_0(M)$ with $\Sigma_p(\mathbb{P})=\sigma_{\mathbb{P},p}$.
Lemma~\ref{L:triangle} implies that $\Sigma_p$ is $1$-Lipschitz.

Specifically, since $p$-Wasserstein distance $W_p$ metrizes weak convergence when $(M,d)$ is compact, we have:
\begin{proposition}
If $(M,d)$ is compact and $\mathbb{P}$ weakly converges to $\mathbb{Q}$, then $\sigma_{\mathbb{P},p}$ converges to $\sigma_{\mathbb{Q},p}$ with respect to $L^\infty$ distance.
\end{proposition}
\begin{remark}\label{rm:faithful}
On the other hand, if $\sigma_{\mathbb{P},p}\equiv\sigma_{\mathbb{Q},p}$, Lemma~\ref{L:dirac} implies $W_p(\mathbb{P},\delta_x)=W_p(\mathbb{Q},\delta_x)$ for any Dirac measure $\delta_x$. Intuitively this means that, at least, $\mathbb{P}$ and $\mathbb{Q}$ look the same from the point of view of all Dirac measures. This implies that $p$-centrality function is a relatively faithful indicator of a probability distribution.
\end{remark}

\subsection{The maximal $p$-centrality discrepancy}

To measure the dissimilarity between two complicated distributions $\mathbb{P}$, $\mathbb{Q}$, we can consider how far the indicators of their push-forwards or projections $f_\ast\mathbb{P}$, $f_\ast\mathbb{Q}$ could fall apart.
According to the dual formulation of $W_1$: 
$$K\cdot W_1(\mathbb{P},\mathbb{Q})=\sup_{f\in Lip(K)}\mathbb{E}_{x\sim f_\ast\mathbb{P}}[x]-\mathbb{E}_{y\sim f_\ast\mathbb{Q}}[y],$$
even considering very simple indicators -- the expectations -- as long as we can search over all $K$-Lipschitz functions $f\in Lip(K)$, we can still approach $W_1$.

Even though a neural network is very powerful on generating all kinds of Lipschitz functions, it may not be able to or have difficulties to generate the optimal push-forward. This may affects the performance of WGAN\cite{Stanczuk2021WassersteinGW}. Hence if we consider more faithful indicators, is it possible to obtain more reliable fake distribution even using sub-optimal push-forward? Motivated by this, we consider Lipschitz functions $f:M\rightarrow\mathbb{R}^n$ and replace the expectations by the $p$-centrality functions. Particularly, for fixed base point $x_0\in\mathbb{R}^n$ we look at discrepancy:
\begin{align*}
L_{p,n,K}(\mathbb{P},\mathbb{Q}):=\sup_{f\in Lip(K)}\sigma_{f_\ast\mathbb{P},p}(x_0)-\sigma_{f_\ast\mathbb{Q},p}(x_0).
\end{align*}

\begin{lemma}
The definition of $L_{p,n,K}$ is independent of the choice of the base point. Or simply $$L_{p,n,K}(\mathbb{P},\mathbb{Q})= \sup_{f\in Lip(K)}\left(\int \|f\|^p d\mathbb{P}\right)^{\frac{1}{p}}-\left(\int \|f\|^p d\mathbb{Q}\right)^{\frac{1}{p}}.$$
\end{lemma}

The following proposition implies that $L_{n,p,K}$ is a direct generalization of Wasserstein distance:
\begin{proposition}\label{P:w1}
If $\operatorname{supp}[\mathbb{P}]$ and $\operatorname{supp}[\mathbb{Q}]$ are both compact, then 
$$L_{1,1,K}(\mathbb{P},\mathbb{Q})=K\cdot W_1(\mathbb{P},\mathbb{Q}).$$
\end{proposition}

Recall that in WGAN, the discriminator is viewed as a $K$-Lipschitz function. 
In our understanding, this requirement is enforced to prevent the discriminator from distorting input distributions too much. More precisely, in the more general setting, the following is true:
\begin{proposition}\label{P:distort}
Given any $K$-Lipschitz map $f:(M,d_M)\rightarrow (N,d_N)$ and Borel probability distributions $\mathbb{P},\mathbb{Q}\in\mathcal{P}(M)$. Then the pushforward distributions $f_\ast\mathbb{P},f_\ast\mathbb{Q}\in\mathcal{P}(N)$ satisfy
$$W_p(f_\ast\mathbb{P},f_\ast\mathbb{Q}) \leq K\cdot W_p(\mathbb{P},\mathbb{Q}).$$
\end{proposition}
More generally, $L_{n,p,K}$ is closely related with $p$-Wasserstein distance:
\begin{theorem}\label{T:main}
For any Borel distributions $\mathbb{P},\mathbb{Q}\in \mathcal{P}(M)$, 
\begin{align*}
L_{p,n,K}(\mathbb{P},\mathbb{Q}) \leq K\cdot W_p(\mathbb{P},\mathbb{Q}).
\end{align*}
\end{theorem}
Also $L_{n,p,K}$ is closely related with an $L^\infty$ distance:
\begin{proposition}\label{P:low}
For any $K$-Lipschitz map $f:M\rightarrow\mathbb{R}^n$,
\begin{align*}
\|\sigma_{f_\ast\mathbb{P},p}-\sigma_{f_\ast\mathbb{Q},p}\|_{\infty}\leq \max\{L_{p,n,K}(\mathbb{P},\mathbb{Q}), L_{p,n,K}(\mathbb{Q},\mathbb{P})\}.
\end{align*}
\end{proposition}

The lower bound in Proposition~\ref{P:low} implies that, when we feed two distributions into the discriminator $f$, as long as some differences retained in the push-forwards $f_\ast\mathbb{P}$ and $f_\ast\mathbb{Q}$, they  would be detected by $L_{p,n,K}$. 
The upper bound in Theorem ~\ref{T:main} implies that, if $\mathbb{P}$ and $\mathbb{Q}$ only differ a little bit under distance $W_p$, then $L_{p,n,K}(\mathbb{P},\mathbb{Q})$ would not change too much. 

As we increase $n$, the $p$-centrality function become more and more faithful which picks up more differences in the discrepancy:
\begin{proposition}\label{P:high-dim}
If integers $n<n'$, then for any $\mathbb{P},\mathbb{Q}\in\mathcal{P}(\mathbb{R}^m)$, we have $L_{p,n,K}(\mathbb{P},\mathbb{Q})\leq L_{p,n',K}(\mathbb{P},\mathbb{Q})$.
\end{proposition}
By Proposition~\ref{P:high-dim} and Theorem~\ref{T:main}, the limit
$$L_{p,K}(\mathbb{P},\mathbb{Q}):=\lim_{n\to \infty} L_{p,n,K}(\mathbb{P},\mathbb{Q})$$ exists and is bounded above by $K\cdot W_p(\mathbb{P},\mathbb{Q})$. Particularly, this bound is tight when $p=1$ (Proposition~\ref{P:w1}).

As a summation, when we use weight regularization such that the discriminator is $K$-Lipschitz and fix some learning rate, using larger critic output dimension $n$ implies that:
\begin{enumerate}
    \item the discriminator may get better approximation of either $L_{p,K}$ or $K\cdot W_p$;
    \item the gradient descent may dive deeper due to larger discrepancy;
    \item the generated fake distribution may be more reliable due to more faithful indicator.
\end{enumerate}

\begin{remark}
Remember that our comparison is under fixed Lipschitz constant $K$.
For example, we can easily scale up the objective function to obtain larger discrepancy, but it is not fair comparison anymore. Because when scaling up objective functions we in fact scaled up both the Lipschitz constant and the maximal possible discrepancy.
\end{remark}









\subsection{Square Root Velocity Transformation}\label{S:srvt}
Section 3.3 suggests us to consider high-dimensional discriminator output. However, if the last layer of discriminator is fully connected, then all output neurons are in symmetric positions and the loss function is permutation invariant. Thus the generalization power of discriminator only depends on the equivalence class obtained by identifying each output vector with its permutations~\cite{badrinarayanan2015understanding,liang2019fisher}. Correspondingly the advantage of high-dimensional output vector would be significantly undermined. 
In order to further improve the performance of our proposed framework, we consider adding an SRVT block to the discriminator to break the symmetric structure. SRVT is usually used in shape analysis to define a distance between curves or functional data. 

Particularly, we view the high-dimensional discriminator output $(x_1,x_2,\cdots,x_n)$ as an ordered sequence.
\begin{definition}
The signed square root function $Q:\mathbb{R}\rightarrow\mathbb{R}$ is given by $Q(x)=\operatorname{sgn}(x)\sqrt{|x|}$.
\end{definition}

Given any differentiable function $f:[0,1]\rightarrow\mathbb{R}$, its SRVT is a function $q:[0,1]\rightarrow\mathbb{R}$ with
\begin{align}
    q:=Q\circ f'=\operatorname{sgn} (f')\sqrt{|f'|}.
\end{align}
SRVT is invertible. Particularly, from $q$ we can recover $f$:
\begin{lemma}
\begin{align}
    f(t) = f(0) + \int_0^t q(s)|q(s)|ds.
\end{align}
\end{lemma}

By assuming $x_0=0$, a discretized SRVT $$S:(x_1,x_2,\cdots,x_n)\in\mathbb{R}^n\mapsto (y_1,\cdots,y_n)\in\mathbb{R}^n$$ is given by 
\begin{align*}
    y_i &= \operatorname{sgn}(x_i-x_{i-1})\sqrt{|x_i-x_{i-1}|}, i=1,2,3,\cdots,n.
\end{align*}

Similarly, $S^{-1}:\mathbb{R}^n\rightarrow\mathbb{R}^n$ is given by
\begin{align*}
    x_i &= \sum_{j=1}^i y_j|y_j|, i=1,2,3,\cdots,n.
\end{align*}

With this transformation, the pullback of $L^2$ norm gives
\begin{align}
    \|(x_1,\cdots,x_n)\|_Q= \sqrt{\sum_{i=1}^n|x_i-x_{i-1}|}
\end{align}

\begin{figure}[ht]
    \centering
    \includegraphics[width=2.8in]{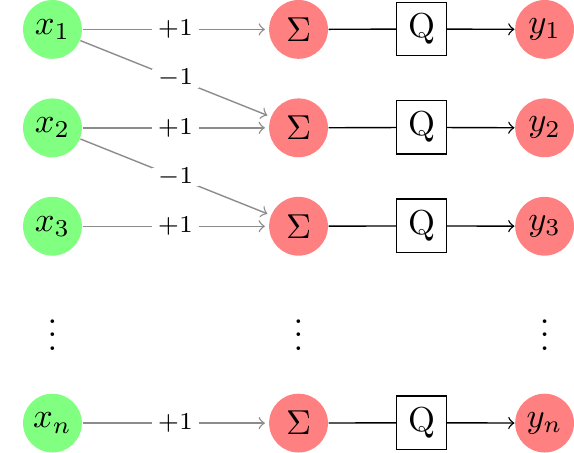}
    \caption{A representation of the SRVT block.}
    \label{fig:sqrt}
\end{figure}

Applying SRVT on a high-dimensional vector results in an ordered sequence which captures the velocity difference at each consecutive position. The discretized SRVT can be represented as a neural network with activation function to be signed square root function $Q$ as depicted in Fig~\ref{fig:sqrt}. Particularly, for the purpose of our paper, each output neuron of SRVT is structurally unique:
\begin{proposition}\label{P:unique}
Any (directed graph) automorphism of the SRVT block leaves each output neuron fixed.
\end{proposition}

Also, the square-root operation has smoothing effect which forces the magnitudes of derivatives to be more concentrated. Thus, values at each output neuron would contribute more similarly to the overall resulting discrepancy. It reduces the risk of over-emphasizing features on certain dimensions and ignoring the rest ones.

\section{Experiments}
In this section we provide experimental results supporting our theoretical analysis and explore various setups to study characteristics of multidimensional critic output. 

\subsection{Datasets and Evaluation Metrics and Implementation Details}
We implemented experiments on CIFAR-10, CIFAR-100 \cite{cifar10}, ImageNet-1K \cite{imagenet}, STL-10 \cite{pmlr-v15-coates11a} and LSUN bedroom \cite{yu15lsun} datasets. 
Results were evaluated with Frechet Inception Distance (FID) \cite{Heusel:2017:GTT:3295222.3295408}, Kernel Inception Distance (KID) \cite{kid} and Precision and Recall (PR) \cite{precision_recall_distributions}. 
For unconditional generation task, we employed StyleGAN2 \cite{Karras2019stylegan2} and ResNet\cite{miyato2018spectral} architectures. 
In StyleGAN2 experiments we followed the default parameter settings provided by \cite{Karras2019stylegan2}.
In ResNet experiments
we used spectral normalization to ensure Lipschitz condition.  
For conditional generation task, we adopted BigGAN \cite{brock2018large} and used their default parameter settings. 
More details can be found in Appendix B.



\subsection{Results}
In the following sections we first present ablation experimental results on CIFAR-10 with analysis, and then report final evaluation scores on all datasets. 

\begin{figure}[ht]
	\centering  
	\includegraphics[width=3.3in,height=2.5in]{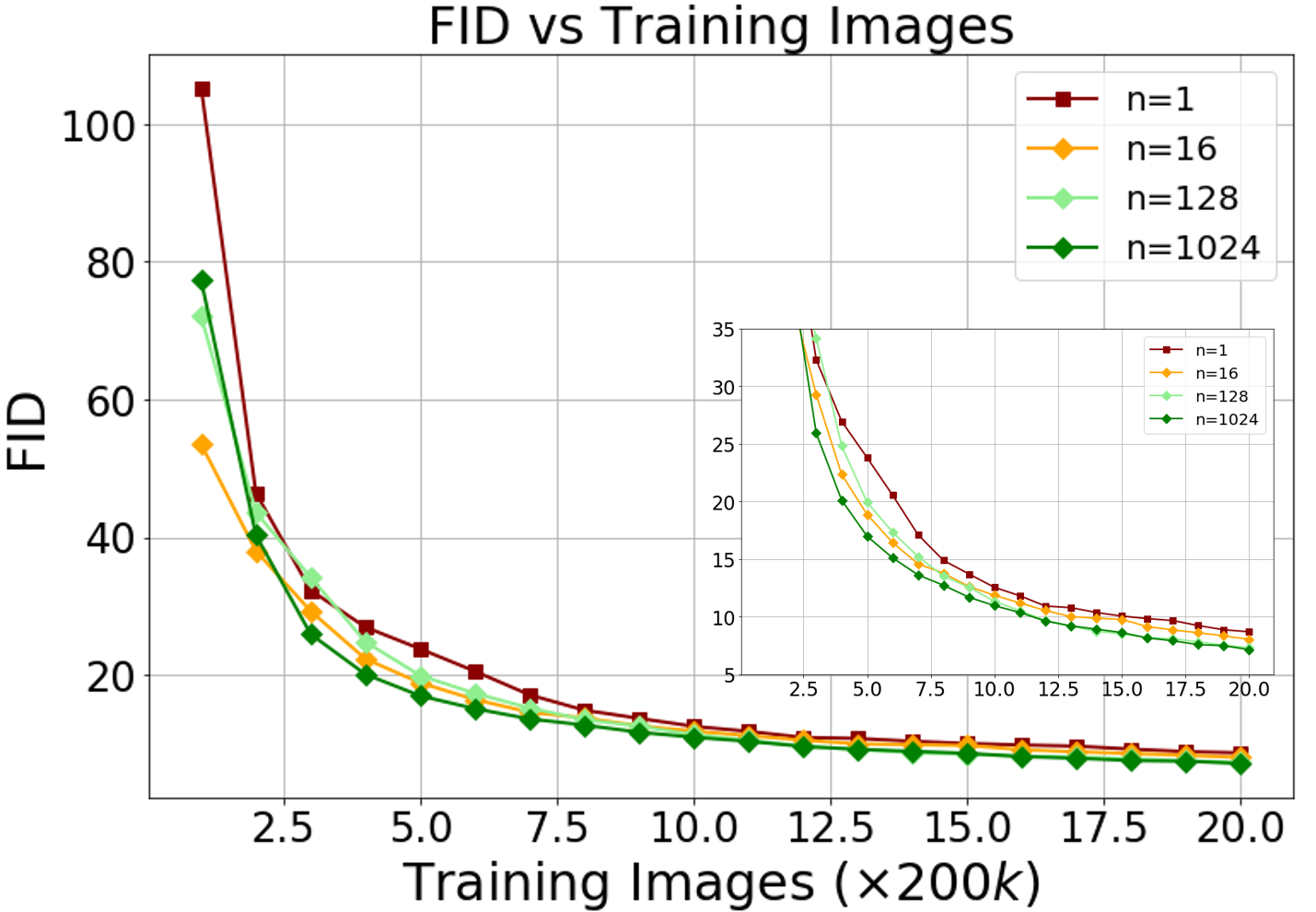}  
	\caption{FID during training on CIFAR-10 with $n=1,16,128$ and $1024$ using StyleGAN2 architectures.} 
	\label{fig:comp_stylegan2}
\end{figure}

\noindent
{\bf Ablation Study:} \\
We first studied the effect of multidimensional critic output using StyleGAN2 network architectures \cite{Karras2019stylegan2}. Figure~\ref{fig:comp_stylegan2} shows recorded FID during training on CIFAR-10. 
Here we applied hinge loss as one common choice for settings with multidimensional output. 
From Figure~\ref{fig:comp_stylegan2} one can see higher $n$ led to faster convergence and consistently competitive results at all training stages.
In training of StyleGAN2, $R_1$ regularization is used as a default choice for regularization.
Note that successful training for higher $n$ in this case requires smaller $\gamma$s. 
Detailed setting for $\gamma$s can be found in Appendix B.

\begin{figure}[ht]
	\centering  
	\includegraphics[width=3.3in,height=2.5in]{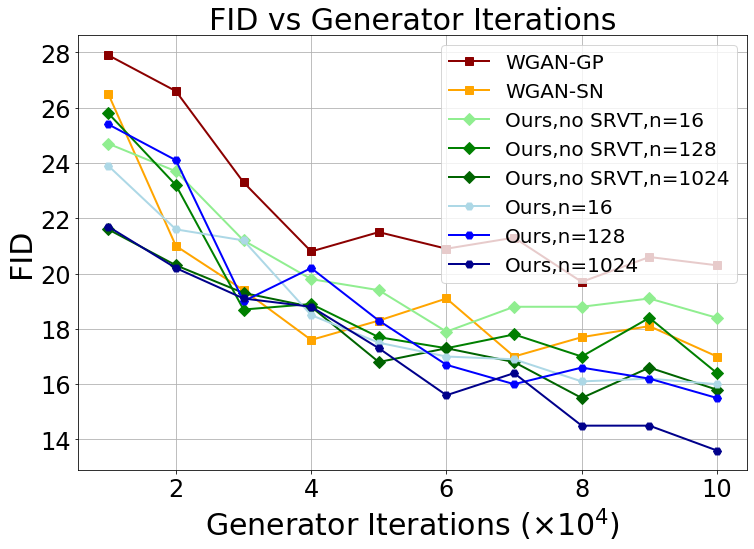}  
	\caption{FID comparison under different settings during training using ResNet backbone.} 
	\label{fig:FID_vs_ite2}
\end{figure}

We then conducted experiments under different settings to explore the effects of $p$-centrality function and SRVT used in our framework. Since our approach is tightly related to WGAN, we also include results from WGAN for comparison. 
In each setting we trained 100K generator iterations on CIFAR-10 using ResNet architectures, and reported average FID scores calculated from 5 runs in Fig~\ref{fig:FID_vs_ite2}. For this experiment we used 10K generated samples for fast evaluation.
One can see without the use of SRVT (three green curves), settings with higher dimensional critic output resulted in better evaluation performances. The pattern is the same when comparing cases with SRVT (three blue curves). These observations are consistent with our Proposition~\ref{P:high-dim}. Furthermore, the results shows the asymmetric transformation boosts performances for different choices of $n$s, especially when $n=1024$ (blue $vs$ green). 
We also provide evaluation results of precision and recall metrics for above settings in Appendix B. 

\begin{figure}[ht]
    \centering
    \begin{tabular}{cc}
    \hspace{-1em}
    {\includegraphics[width=.24\textwidth]{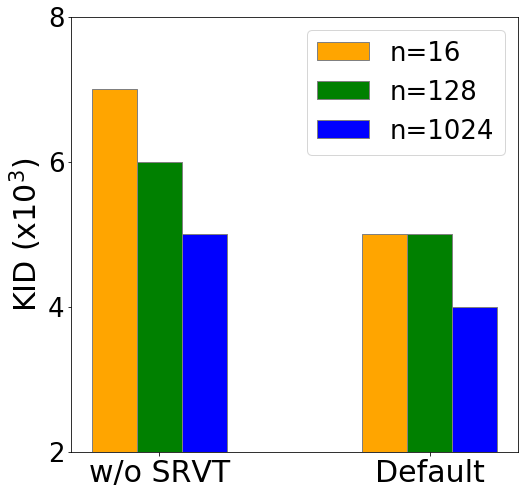}}\hspace{-1em} & 
    {\includegraphics[width=.245\textwidth]{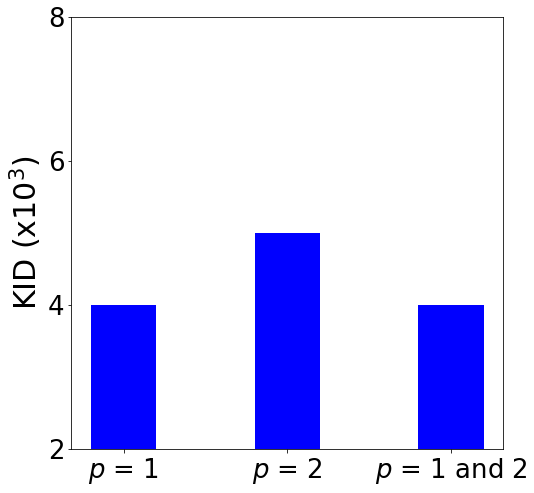}} \\
    (a)\hspace{-1em} & (b) 
    \end{tabular}
    \caption{KID evaluation under different settings. (a) Left: without SRVT; Right: default setting with SRVT. (b) Evaluation with SRVT under different $p$s with fixed $n=1024$.}
    \label{fig:comp_KID}
\end{figure}

We also present comparisons using KID under different settings in Fig~\ref{fig:comp_KID}. Results in Fig~\ref{fig:comp_KID}(a) are aligned with previous evaluations which shows the advantage of using higher dimensional critic output. Performance was further boosted with SRVT.  Fig~\ref{fig:comp_KID}(b) shows KID evaluations under different choices of $p$s, where SRVT was used with fixed $n=1024$. We observe using $p=1$ only, or both $p=1$ and $2$ resulted in better performance compared with using $p=2$ only. 
We also explored involving higher $p$ ($p > 2$) yet did not see improvement in terms of evaluations.
In the following we used $p = 1$ as the default setting. 
 

In the following we display our final evaluation results. For fair comparison we list comparable results using the same network architectures.




\noindent
{\bf Quantitative Results:}\\
To compare GAN objectives, we present evaluations of FID on unconditional generation experiments averaged over 5 random runs in Table~\ref{tab:unconditional} . We compare with methods related to our work, including WGAN-GP \cite{gulrajani2017improved}, MMD GAN-rq \cite{li2017mmd}, SNGAN  \cite{miyato2018spectral}, CTGAN \cite{wei2018improving}, Sphere GAN \cite{Park_2019_CVPR}, SWGAN \cite{Wu_2019_CVPR}, CRGAN \cite {Zhang2020Consistency} and DGflow \cite{ansari2021refining}. 

\begin{table}[!ht]
	\centering
	\caption{FIDs$(\downarrow)$ from unconditional generation experiments on CIFAR-10 with ResNet architectures.}
	\label{tab:unconditional}
	\begin{tabular}{l|ccc}
		\hline
		Method  & CIFAR-10 & STL-10 & LSUN\\
		\hline
		WGAN-GP  & 19.0(0.8) & 55.1 & 26.9(1.1)\\
		SNGAN & 14.1(0.6) & 40.1(0.5) & 31.3(2.1) \\
		MMD GAN-rq & - & - & 32.0 \\
		CTGAN & 17.6(0.7) & - & 19.5(1.2) \\
		Sphere GAN  & 17.1 & 31.4 & 16.9 \\
		SWGAN  & 17.0(1.0) & - & 14.9(1.0)\\
		CRGAN & 14.6 & - & -\\
		DGflow & 9.6(0.1) & - & - \\
		
		\hline
		Ours   & {8.5(0.3)} & {26.1(0.4)} & {14.2(0.2)}\\
		\hline
	\end{tabular}
\end{table}

Here we also present evaluation results of unconditional experiments on ImageNet using StyleGAN2 architectures.
Table~\ref{tab:stylegan2-cifar-imagenet} shows the feasibility of using high-dimensional critic output in large-scale settings.
With comparable FIDs, the precision-recall scores indicate that high-dimensional critic output potentially improves diversity of results. 

\begin{table}[!ht]
  \centering
  \begin{tabular}{c|ccc}
    \toprule
    $n$ & FID & Precision & Recall\\
    \midrule
    $1$ &55.82 &0.677 &0.883\\
    $1024$ &53.66  &0.637 &0.901\\
    \bottomrule
  \end{tabular}
  \caption{Evaluations on 256 $\times$ 256 ImageNet experiments with $n=1$ and $1024$ using StyleGAN2 architectures.}
  \label{tab:stylegan2-cifar-imagenet}
\end{table}

\begin{table}[!ht]
	\centering
	\caption{FIDs$(\downarrow)$ from conditional generation experiments with BigGAN architectures.}
	\label{tab:conditional}
	\begin{tabular}{c|cc}
		\hline
		Objective  & CIFAR-10 & CIFAR-100\\
		\hline
		Hinge & 9.7(0.1) & 13.6(0.1)\\
		Ours   & 8.9(0.1) & 12.3(0.1) \\
		\hline
	\end{tabular}
\end{table}

As presented in Table~\ref{tab:unconditional}, the proposed method led to competitive results in comparable settings on the three datasets.

For conditional generation, we show evaluation results from the original BigGAN setting and the proposed objective in Table~\ref{tab:conditional}. The results indicate the proposed framework can also be applied in the more sophisticated training setting and obtain competitive performance.


\section{Conclusion and Discussion}
In this paper we have explored the properties of multiple critic outputs in GANs 
based on the proposed the {\it maximal $p$-centrality discrepancy}. 
We have further introduced an asymmetrical (square-root velocity) transformation added to discriminator to break the symmetric structure of its network output. The use of the nonparametric transformation takes advantage of multidimensional features and improves the generalization capability of the network.
Although the properties are investigated in a WGAN framework, the general pattern can also be extended to other frameworks which utilize min-max discrepancy as objectives.


\bibliography{awgan}

\begin{thebibliography}{55}
\providecommand{\natexlab}[1]{#1}
\providecommand{\url}[1]{\texttt{#1}}
\expandafter\ifx\csname urlstyle\endcsname\relax
  \providecommand{\doi}[1]{doi: #1}\else
  \providecommand{\doi}{doi: \begingroup \urlstyle{rm}\Url}\fi

\bibitem[Ansari et~al.(2020)Ansari, Scarlett, and Soh]{Ansari_2020_CVPR}
Ansari, A.~F., Scarlett, J., and Soh, H.
\newblock A characteristic function approach to deep implicit generative
  modeling.
\newblock In \emph{Proceedings of the IEEE/CVF Conference on Computer Vision
  and Pattern Recognition (CVPR)}, June 2020.

\bibitem[Ansari et~al.(2021)Ansari, Ang, and Soh]{ansari2021refining}
Ansari, A.~F., Ang, M.~L., and Soh, H.
\newblock Refining deep generative models via discriminator gradient flow.
\newblock In \emph{International Conference on Learning Representations}, 2021.

\bibitem[Arjovsky \& Bottou(2017)Arjovsky and Bottou]{arjovsky2017principled}
Arjovsky, M. and Bottou, L.
\newblock Towards principled methods for training generative adversarial
  networks, 2017.

\bibitem[Arjovsky et~al.(2017)Arjovsky, Chintala, and Bottou]{wgan2017}
Arjovsky, M., Chintala, S., and Bottou, L.
\newblock {W}asserstein generative adversarial networks.
\newblock In \emph{Proceedings of the 34th International Conference on Machine
  Learning}, volume~70, pp.\  214--223, 2017.

\bibitem[Arnaudon et~al.(2013)Arnaudon, Barbaresco, and
  Yang]{arnaudon2013medians}
Arnaudon, M., Barbaresco, F., and Yang, L.
\newblock Medians and means in riemannian geometry: existence, uniqueness and
  computation.
\newblock In \emph{Matrix Information Geometry}, pp.\  169--197. Springer,
  2013.

\bibitem[Badrinarayanan et~al.(2015)Badrinarayanan, Mishra, and
  Cipolla]{badrinarayanan2015understanding}
Badrinarayanan, V., Mishra, B., and Cipolla, R.
\newblock Understanding symmetries in deep networks.
\newblock \emph{arXiv preprint arXiv:1511.01029}, 2015.

\bibitem[Bhattacharya \& Patrangenaru(2003)Bhattacharya and
  Patrangenaru]{bhattacharya2003large}
Bhattacharya, R. and Patrangenaru, V.
\newblock Large sample theory of intrinsic and extrinsic sample means on
  manifolds.
\newblock \emph{The Annals of Statistics}, 31\penalty0 (1):\penalty0 1--29,
  2003.

\bibitem[Bińkowski et~al.(2018{\natexlab{a}})Bińkowski, Sutherland, Arbel,
  and Gretton]{2018demystifying}
Bińkowski, M., Sutherland, D.~J., Arbel, M., and Gretton, A.
\newblock Demystifying {MMD} {GAN}s.
\newblock In \emph{International Conference on Learning Representations},
  2018{\natexlab{a}}.

\bibitem[Bińkowski et~al.(2018{\natexlab{b}})Bińkowski, Sutherland, Arbel,
  and Gretton]{kid}
Bińkowski, M., Sutherland, D.~J., Arbel, M., and Gretton, A.
\newblock Demystifying {MMD} {GAN}s.
\newblock In \emph{International Conference on Learning Representations},
  2018{\natexlab{b}}.

\bibitem[Brock et~al.(2019)Brock, Donahue, and Simonyan]{brock2018large}
Brock, A., Donahue, J., and Simonyan, K.
\newblock Large scale {GAN} training for high fidelity natural image synthesis.
\newblock In \emph{International Conference on Learning Representations}, 2019.

\bibitem[Chan et~al.(2021)Chan, Monteiro, Kellnhofer, Wu, and
  Wetzstein]{Chan_2021_CVPR}
Chan, E.~R., Monteiro, M., Kellnhofer, P., Wu, J., and Wetzstein, G.
\newblock Pi-gan: Periodic implicit generative adversarial networks for
  3d-aware image synthesis.
\newblock In \emph{Proceedings of the IEEE/CVF Conference on Computer Vision
  and Pattern Recognition (CVPR)}, pp.\  5799--5809, June 2021.

\bibitem[Coates et~al.(2011)Coates, Ng, and Lee]{pmlr-v15-coates11a}
Coates, A., Ng, A., and Lee, H.
\newblock An analysis of single-layer networks in unsupervised feature
  learning.
\newblock In \emph{Proceedings of the Fourteenth International Conference on
  Artificial Intelligence and Statistics}, volume~15, pp.\  215--223, 2011.

\bibitem[Deng et~al.(2009)Deng, Dong, Socher, Li, Li, and Fei-Fei]{imagenet}
Deng, J., Dong, W., Socher, R., Li, L.-J., Li, K., and Fei-Fei, L.
\newblock Imagenet: A large-scale hierarchical image database.
\newblock In \emph{2009 IEEE Conference on Computer Vision and Pattern
  Recognition}, pp.\  248--255, 2009.

\bibitem[Deshpande et~al.(2018)Deshpande, Zhang, and
  Schwing]{Deshpande_2018_CVPR}
Deshpande, I., Zhang, Z., and Schwing, A.~G.
\newblock Generative modeling using the sliced wasserstein distance.
\newblock In \emph{Proceedings of the IEEE Conference on Computer Vision and
  Pattern Recognition (CVPR)}, June 2018.

\bibitem[Deshpande et~al.(2019)Deshpande, Hu, Sun, Pyrros, Siddiqui, Koyejo,
  Zhao, Forsyth, and Schwing]{Deshpande_2019_CVPR}
Deshpande, I., Hu, Y.-T., Sun, R., Pyrros, A., Siddiqui, N., Koyejo, S., Zhao,
  Z., Forsyth, D., and Schwing, A.~G.
\newblock Max-sliced wasserstein distance and its use for gans.
\newblock In \emph{Proceedings of the IEEE/CVF Conference on Computer Vision
  and Pattern Recognition (CVPR)}, June 2019.

\bibitem[Dziugaite et~al.(2015)Dziugaite, Roy, and
  Ghahramani]{dziugaite2015training}
Dziugaite, G.~K., Roy, D.~M., and Ghahramani, Z.
\newblock Training generative neural networks via maximum mean discrepancy
  optimization.
\newblock In \emph{Proceedings of the Thirty-First Conference on Uncertainty in
  Artificial Intelligence}, pp.\  258--267, 2015.

\bibitem[Glorot \& Bengio(2010)Glorot and Bengio]{glorot2010understanding}
Glorot, X. and Bengio, Y.
\newblock Understanding the difficulty of training deep feedforward neural
  networks.
\newblock In \emph{Proceedings of the thirteenth international conference on
  artificial intelligence and statistics}, pp.\  249--256, 2010.

\bibitem[Goodfellow et~al.(2014)Goodfellow, Pouget-Abadie, Mirza, Xu,
  Warde-Farley, Ozair, Courville, and Bengio]{goodfellow2014generative}
Goodfellow, I., Pouget-Abadie, J., Mirza, M., Xu, B., Warde-Farley, D., Ozair,
  S., Courville, A., and Bengio, Y.
\newblock Generative adversarial nets.
\newblock In \emph{Advances in neural information processing systems}, pp.\
  2672--2680, 2014.

\bibitem[Grove \& Karcher(1973)Grove and Karcher]{grove1973conjugatec}
Grove, K. and Karcher, H.
\newblock How to conjugatec ${C}^1$-close group actions.
\newblock \emph{Mathematische Zeitschrift}, 132\penalty0 (1):\penalty0 11--20,
  1973.

\bibitem[Gulrajani et~al.(2017)Gulrajani, Ahmed, Arjovsky, Dumoulin, and
  Courville]{gulrajani2017improved}
Gulrajani, I., Ahmed, F., Arjovsky, M., Dumoulin, V., and Courville, A.
\newblock Improved training of wasserstein gans.
\newblock In \emph{Proceedings of the 31st International Conference on Neural
  Information Processing Systems}, NIPS'17, pp.\  5769--5779, 2017.

\bibitem[Han et~al.(2021)Han, Chen, and Liu]{Han_Chen_Liu_2021}
Han, X., Chen, X., and Liu, L.-P.
\newblock Gan ensemble for anomaly detection.
\newblock \emph{Proceedings of the AAAI Conference on Artificial Intelligence},
  35\penalty0 (5):\penalty0 4090--4097, May 2021.

\bibitem[Hang et~al.(2019)Hang, M{\'e}moli, and Mio]{hang2019topological}
Hang, H., M{\'e}moli, F., and Mio, W.
\newblock A topological study of functional data and fr{\'e}chet functions of
  metric measure spaces.
\newblock \emph{Journal of Applied and Computational Topology}, 3\penalty0
  (4):\penalty0 359--380, 2019.

\bibitem[He et~al.(2015)He, Zhang, Ren, and Sun]{he2015delving}
He, K., Zhang, X., Ren, S., and Sun, J.
\newblock Delving deep into rectifiers: Surpassing human-level performance on
  imagenet classification.
\newblock In \emph{Proceedings of the IEEE international conference on computer
  vision}, pp.\  1026--1034, 2015.

\bibitem[He et~al.(2016{\natexlab{a}})He, Zhang, Ren, and Sun]{he2016deep}
He, K., Zhang, X., Ren, S., and Sun, J.
\newblock Deep residual learning for image recognition.
\newblock In \emph{Proceedings of the IEEE conference on computer vision and
  pattern recognition}, pp.\  770--778, 2016{\natexlab{a}}.

\bibitem[He et~al.(2016{\natexlab{b}})He, Zhang, Ren, and Sun]{he2016identity}
He, K., Zhang, X., Ren, S., and Sun, J.
\newblock Identity mappings in deep residual networks.
\newblock In \emph{European conference on computer vision}, pp.\  630--645.
  Springer, 2016{\natexlab{b}}.

\bibitem[Heim(2019)]{Heim_2019_CVPR}
Heim, E.
\newblock Constrained generative adversarial networks for interactive image
  generation.
\newblock In \emph{The IEEE Conference on Computer Vision and Pattern
  Recognition (CVPR)}, June 2019.

\bibitem[Heusel et~al.(2017)Heusel, Ramsauer, Unterthiner, Nessler, and
  Hochreiter]{Heusel:2017:GTT:3295222.3295408}
Heusel, M., Ramsauer, H., Unterthiner, T., Nessler, B., and Hochreiter, S.
\newblock Gans trained by a two time-scale update rule converge to a local nash
  equilibrium.
\newblock In \emph{Proceedings of the 31st International Conference on Neural
  Information Processing Systems}, NIPS'17, pp.\  6629--6640, 2017.

\bibitem[Huang et~al.(2017)Huang, Liu, Van Der~Maaten, and
  Weinberger]{huang2017densely}
Huang, G., Liu, Z., Van Der~Maaten, L., and Weinberger, K.~Q.
\newblock Densely connected convolutional networks.
\newblock In \emph{Proceedings of the IEEE conference on computer vision and
  pattern recognition}, pp.\  4700--4708, 2017.

\bibitem[Karras et~al.(2020{\natexlab{a}})Karras, Aittala, Hellsten, Laine,
  Lehtinen, and Aila]{Karras2020ada}
Karras, T., Aittala, M., Hellsten, J., Laine, S., Lehtinen, J., and Aila, T.
\newblock Training generative adversarial networks with limited data.
\newblock In \emph{Proc. NeurIPS}, 2020{\natexlab{a}}.

\bibitem[Karras et~al.(2020{\natexlab{b}})Karras, Laine, Aittala, Hellsten,
  Lehtinen, and Aila]{Karras2019stylegan2}
Karras, T., Laine, S., Aittala, M., Hellsten, J., Lehtinen, J., and Aila, T.
\newblock Analyzing and improving the image quality of {StyleGAN}.
\newblock In \emph{Proc. CVPR}, 2020{\natexlab{b}}.

\bibitem[Kolouri et~al.(2016)Kolouri, Zou, and Rohde]{kolouri2016sliced}
Kolouri, S., Zou, Y., and Rohde, G.~K.
\newblock Sliced wasserstein kernels for probability distributions.
\newblock In \emph{Proceedings of the IEEE Conference on Computer Vision and
  Pattern Recognition}, pp.\  5258--5267, 2016.

\bibitem[Kolouri et~al.(2019)Kolouri, Nadjahi, Simsekli, Badeau, and
  Rohde]{NEURIPS2019_f0935e4c}
Kolouri, S., Nadjahi, K., Simsekli, U., Badeau, R., and Rohde, G.
\newblock Generalized sliced wasserstein distances.
\newblock In \emph{Advances in Neural Information Processing Systems},
  volume~32, pp.\  261--272, 2019.

\bibitem[Krizhevsky et~al.(2010)Krizhevsky, Nair, and Hinton]{cifar10}
Krizhevsky, A., Nair, V., and Hinton, G.
\newblock Cifar-10 (canadian institute for advanced research).
\newblock \emph{URL http://www. cs. toronto. edu/kriz/cifar. html}, 5, 2010.

\bibitem[Lee et~al.(2019)Lee, Batra, Baig, and Ulbricht]{Lee_2019_CVPR}
Lee, C.-Y., Batra, T., Baig, M.~H., and Ulbricht, D.
\newblock Sliced wasserstein discrepancy for unsupervised domain adaptation.
\newblock In \emph{Proceedings of the IEEE/CVF Conference on Computer Vision
  and Pattern Recognition (CVPR)}, June 2019.

\bibitem[Li et~al.(2017)Li, Chang, Cheng, Yang, and P{\'o}czos]{li2017mmd}
Li, C.-L., Chang, W.-C., Cheng, Y., Yang, Y., and P{\'o}czos, B.
\newblock Mmd gan: Towards deeper understanding of moment matching network.
\newblock In \emph{Advances in Neural Information Processing Systems}, pp.\
  2203--2213, 2017.

\bibitem[Li et~al.(2015)Li, Swersky, and Zemel]{gmmn2015}
Li, Y., Swersky, K., and Zemel, R.~S.
\newblock Generative moment matching networks.
\newblock \emph{CoRR}, abs/1502.02761, 2015.

\bibitem[Liang et~al.(2019)Liang, Poggio, Rakhlin, and Stokes]{liang2019fisher}
Liang, T., Poggio, T., Rakhlin, A., and Stokes, J.
\newblock Fisher-rao metric, geometry, and complexity of neural networks.
\newblock In \emph{The 22nd International Conference on Artificial Intelligence
  and Statistics}, pp.\  888--896. PMLR, 2019.

\bibitem[Miyato et~al.(2018)Miyato, Kataoka, Koyama, and
  Yoshida]{miyato2018spectral}
Miyato, T., Kataoka, T., Koyama, M., and Yoshida, Y.
\newblock Spectral normalization for generative adversarial networks.
\newblock In \emph{International Conference on Learning Representations}, 2018.

\bibitem[Mroueh \& Sercu(2017)Mroueh and Sercu]{mroueh2017fisher}
Mroueh, Y. and Sercu, T.
\newblock Fisher gan.
\newblock In \emph{Advances in Neural Information Processing Systems}, pp.\
  2513--2523, 2017.

\bibitem[Mroueh et~al.(2017{\natexlab{a}})Mroueh, Li, Sercu, Raj, and
  Cheng]{sobolevgan2017}
Mroueh, Y., Li, C., Sercu, T., Raj, A., and Cheng, Y.
\newblock Sobolev {GAN}.
\newblock \emph{CoRR}, abs/1711.04894, 2017{\natexlab{a}}.

\bibitem[Mroueh et~al.(2017{\natexlab{b}})Mroueh, Sercu, and Goel]{mcgan2017}
Mroueh, Y., Sercu, T., and Goel, V.
\newblock {M}c{G}an: Mean and covariance feature matching {GAN}.
\newblock In \emph{Proceedings of the 34th International Conference on Machine
  Learning}, volume~70, pp.\  2527--2535, 2017{\natexlab{b}}.

\bibitem[Nauata et~al.(2020)Nauata, Chang, Cheng, Mori, and
  Furukawa]{nauata2020house}
Nauata, N., Chang, K.-H., Cheng, C.-Y., Mori, G., and Furukawa, Y.
\newblock House-gan: Relational generative adversarial networks for
  graph-constrained house layout generation.
\newblock In \emph{European Conference on Computer Vision}, pp.\  162--177.
  Springer, 2020.

\bibitem[Niemeyer \& Geiger(2021)Niemeyer and Geiger]{Niemeyer_2021_CVPR}
Niemeyer, M. and Geiger, A.
\newblock Giraffe: Representing scenes as compositional generative neural
  feature fields.
\newblock In \emph{Proceedings of the IEEE/CVF Conference on Computer Vision
  and Pattern Recognition (CVPR)}, pp.\  11453--11464, June 2021.

\bibitem[Park \& Kwon(2019)Park and Kwon]{Park_2019_CVPR}
Park, S.~W. and Kwon, J.
\newblock Sphere generative adversarial network based on geometric moment
  matching.
\newblock In \emph{The IEEE Conference on Computer Vision and Pattern
  Recognition (CVPR)}, June 2019.

\bibitem[Rabin et~al.(2011)Rabin, Peyr{\'e}, Delon, and
  Bernot]{rabin2011wasserstein}
Rabin, J., Peyr{\'e}, G., Delon, J., and Bernot, M.
\newblock Wasserstein barycenter and its application to texture mixing.
\newblock In \emph{International Conference on Scale Space and Variational
  Methods in Computer Vision}, pp.\  435--446. Springer, 2011.

\bibitem[Sajjadi et~al.(2018)Sajjadi, Bachem, Lu{\v c}i{\'c}, Bousquet, and
  Gelly]{precision_recall_distributions}
Sajjadi, M.~S.~M., Bachem, O., Lu{\v c}i{\'c}, M., Bousquet, O., and Gelly, S.
\newblock {Assessing Generative Models via Precision and Recall}.
\newblock In \emph{{Advances in Neural Information Processing Systems
  (NeurIPS)}}, 2018.

\bibitem[Srivastava \& Klassen(2016)Srivastava and
  Klassen]{srivastava2016functional}
Srivastava, A. and Klassen, E.~P.
\newblock \emph{Functional and shape data analysis}, volume~1.
\newblock Springer, 2016.

\bibitem[{Srivastava} et~al.(2011){Srivastava}, {Klassen}, {Joshi}, and
  {Jermyn}]{5601739}
{Srivastava}, A., {Klassen}, E., {Joshi}, S.~H., and {Jermyn}, I.~H.
\newblock Shape analysis of elastic curves in euclidean spaces.
\newblock \emph{IEEE Transactions on Pattern Analysis and Machine
  Intelligence}, 33\penalty0 (7):\penalty0 1415--1428, July 2011.

\bibitem[Stanczuk et~al.(2021)Stanczuk, Etmann, Kreusser, and
  Sch{\"o}nlieb]{Stanczuk2021WassersteinGW}
Stanczuk, J., Etmann, C., Kreusser, L.~M., and Sch{\"o}nlieb, C.-B.
\newblock Wasserstein gans work because they fail (to approximate the
  wasserstein distance).
\newblock \emph{ArXiv}, abs/2103.01678, 2021.

\bibitem[Wei et~al.(2018)Wei, Liu, Wang, and Gong]{wei2018improving}
Wei, X., Liu, Z., Wang, L., and Gong, B.
\newblock Improving the improved training of wasserstein {GAN}s.
\newblock In \emph{International Conference on Learning Representations}, 2018.

\bibitem[Wu et~al.(2019)Wu, Huang, Acharya, Li, Thoma, Paudel, and
  Gool]{Wu_2019_CVPR}
Wu, J., Huang, Z., Acharya, D., Li, W., Thoma, J., Paudel, D.~P., and Gool,
  L.~V.
\newblock Sliced wasserstein generative models.
\newblock In \emph{Proceedings of the IEEE/CVF Conference on Computer Vision
  and Pattern Recognition (CVPR)}, June 2019.

\bibitem[Yang et~al.(2022)Yang, Wen, and
  Davatzikos]{yang2022surrealgansemisupervised}
Yang, Z., Wen, J., and Davatzikos, C.
\newblock Surreal-{GAN}:semi-supervised representation learning via {GAN} for
  uncovering heterogeneous disease-related imaging patterns.
\newblock In \emph{International Conference on Learning Representations}, 2022.

\bibitem[Yu et~al.(2015)Yu, Zhang, Song, Seff, and Xiao]{yu15lsun}
Yu, F., Zhang, Y., Song, S., Seff, A., and Xiao, J.
\newblock Lsun: Construction of a large-scale image dataset using deep learning
  with humans in the loop.
\newblock \emph{arXiv preprint arXiv:1506.03365}, 2015.

\bibitem[Yu et~al.(2022)Yu, Tack, Mo, Kim, Kim, Ha, and Shin]{yu2022generating}
Yu, S., Tack, J., Mo, S., Kim, H., Kim, J., Ha, J.-W., and Shin, J.
\newblock Generating videos with dynamics-aware implicit generative adversarial
  networks.
\newblock In \emph{International Conference on Learning Representations}, 2022.

\bibitem[Zhang et~al.(2020)Zhang, Zhang, Odena, and Lee]{Zhang2020Consistency}
Zhang, H., Zhang, Z., Odena, A., and Lee, H.
\newblock Consistency regularization for generative adversarial networks.
\newblock In \emph{International Conference on Learning Representations}, 2020.

\end{thebibliography}
\bibliographystyle{icml2022}

\newpage 

\newpage

\begin{appendices}

\section{Proofs}
\subsection{Lemma 3.3}
\begin{proof}
For any $x\in M$, by Lemma 3.2 and triangle inequality we have
\begin{align*}
    |\sigma_{\mathbb{P},p}(x)-\sigma_{\mathbb{Q},p}(x)|\leq W_p(\mathbb{P},\mathbb{Q})\leq |\sigma_{\mathbb{P},p}(x)+\sigma_{\mathbb{Q},p}(x)|.
\end{align*}
The result follows by letting $x$ run over all $M$.
\end{proof}

\subsection{Lemma 3.6}
\begin{proof}
Let $\phi$ be the translation map on $\mathbb{R}^n$ with $\phi(y)=y+x_0$. Then $g:=\phi^{-1}\circ f\in Lip(K)$ iff. $f\in Lip(K)$ and
\begin{align*}
& L_{p,n,K}(\mathbb{P},\mathbb{Q}) =  \sup_{f\in Lip(K)}\sigma_{f_\ast\mathbb{P},p}(\phi(0))-\sigma_{f_\ast\mathbb{Q},p}(\phi(0))\\
& = \sup_{f\in Lip(K)}\sigma_{(\phi^{-1}\circ f)_\ast\mathbb{P},p}(0)-\sigma_{(\phi^{-1}\circ f)_\ast\mathbb{Q},p}(0)\\
& = \sup_{g\in Lip(K)}\sigma_{g_\ast\mathbb{P},p}(0)-\sigma_{g_\ast\mathbb{Q},p}(0)\\
& = \sup_{f\in Lip(K)}\left(\int \|f\|^p d\mathbb{P}\right)^{\frac{1}{p}}-\left(\int \|f\|^p d\mathbb{Q}\right)^{\frac{1}{p}}
.
\end{align*}
\end{proof}

\subsection{Proposition 3.7}
\begin{proof}
Since $f\in Lip(K)$ implies $|f|\in Lip(K)$, we easily have $L_{1,1,K}\leq K\cdot W_1$.

On the other hand, for any $\epsilon>0$, there exists a $K$-Lipschitz map $f:M\rightarrow\mathbb{R}$ s.t. $\int f d\mathbb{P}-\int f d\mathbb{Q}>K\cdot W_1(\mathbb{P},\mathbb{Q})-\epsilon$. Let $D = \operatorname{supp}[\mathbb{P}]\cup\operatorname{supp}[\mathbb{Q}]$ and $c = \min_{x\in D}f(x)$, then $f-c\geq 0$ and $\int f d\mathbb{P}-\int f d\mathbb{Q}=\int (f-c) d\mathbb{P}-\int (f-c) d\mathbb{Q}=\int |f-c| d\mathbb{P}-\int |f-c| d\mathbb{Q}\leq L_{1,1,K}(\mathbb{P},\mathbb{Q})$. Hence $L_{1,1,K}(\mathbb{P},\mathbb{Q})\geq K\cdot W_1(\mathbb{P},\mathbb{Q})-\epsilon$ for any $\epsilon>0$ which implies $L_{1,1,K}\geq K\cdot W_1$.
\end{proof}

\subsection{Proposition 3.8}
\begin{proof}
Let $\Gamma(\mathbb{P},\mathbb{Q})$ be the set of all joint probability measures of $\mathbb{P}$ and $\mathbb{Q}$. For any $\gamma\in\Gamma(\mathbb{P},\mathbb{Q})$, we have $f_\ast\gamma\in \Gamma(f_\ast\mathbb{P},f_\ast\mathbb{Q})$. By definition of the $p$-Wasserstein distance,
\begin{align*}
& W_p(f_\ast\mathbb{P},f_\ast\mathbb{Q}) \\
= & \inf_{\gamma'\in\Gamma(f_\ast\mathbb{P},f_\ast\mathbb{Q})}\left(\int_{N\times N}d_N^p(y_1,y_2)d\gamma'(y_1,y_2)\right)^{1/p} \\
\leq & \inf_{\gamma\in\Gamma(\mathbb{P},\mathbb{Q})}\left(\int_{N\times N}d_N^p(y_1,y_2)d(f_\ast\gamma)(y_1,y_2)\right)^{1/p} \\
= & \inf_{\gamma\in\Gamma(\mathbb{P},\mathbb{Q})} \left(\int_{M\times M} d^p_N(f(x_1),f(x_2)) d\gamma(x_1,x_2)\right)^{1/p} \\
\leq & \inf_{\gamma\in\Gamma(\mathbb{P},\mathbb{Q})}  \left(\int_{M\times M} K^p\cdot d^p_M(x_1,x_2) d\gamma(x_1,x_2)\right)^{1/p} \\
= & K\cdot \inf_{\gamma\in\Gamma(\mathbb{P},\mathbb{Q})} \left(\int_{M\times M} d^p_M(y_1,y_2) d\gamma(y_1,y_2)\right)^{1/p}\\
= & K\cdot W_p(\mathbb{P},\mathbb{Q}).
\end{align*}
\end{proof}

\subsection{Theorem 3.9}
\begin{proof}
By Lemma 3.2, we have $$L_{p,n,K}(\mathbb{P},\mathbb{Q})=\sup_{f\in Lip(K)}W_p(f_\ast\mathbb{P},\delta_0)-W_p(f_\ast\mathbb{Q},\delta_0).$$
Applying triangle inequality and Proposition~\ref{P:distort}, we have 
$$L_{p,n,K}(\mathbb{P},\mathbb{Q})\leq \sup_{f\in Lip(K)}W_p(f_\ast\mathbb{P},f_\ast\mathbb{Q})\leq K\cdot W_p(\mathbb{P},\mathbb{Q}).$$
\end{proof}

\subsection{Proposition 3.10}
\begin{proof}
\begin{align*}
& \|\sigma_{f_\ast\mathbb{P},p}-\sigma_{f_\ast\mathbb{Q},p}\|_\infty\\
= & \sup_{x_0\in\mathbb{R}^n}\big|\sigma_{f_\ast\mathbb{P},p}(x_0)-\sigma_{f_\ast\mathbb{Q},p}(x_0)\big|\\
\leq & \sup_{x_0\in\mathbb{R}^n}\sup_{f\in Lip(K)}\big|\sigma_{f_\ast\mathbb{P},p}(x_0)-\sigma_{f_\ast\mathbb{Q},p}(x_0)\big|\\
= & \sup_{x_0\in\mathbb{R}^n} \max\{L_{p,n,K}(\mathbb{P},\mathbb{Q}),L_{p,n,K}(\mathbb{Q},\mathbb{P})\}.
\end{align*}
\end{proof}

\subsection{Proposition 3.11}
\begin{proof}
For any $n<n'$ we have natural embedding $\mathbb{R}^n\hookrightarrow\mathbb{R}^{n'}$. Hence any $K$-Lipschitz function with domain  $\mathbb{R}^n$ can also be viewed as a $K$-Lipschitz function with domain $\mathbb{R}^{n'}$. Hence larger $n$ gives larger candidate pool for searching the maximal discrepancy and the result follows.
\end{proof}

\subsection{Proposition 3.15}
\begin{proof}
View the SRVT block as a directed graph, then all output neurons has out-degree $0$. By the definition of discritized SRVT, there is a unique output neuron $v_0$ with in-degree $1$ and any two different output neurons have different distance to $v_0$. Since any automorphism of directed graph would preserve in-degrees, out-degrees and distance, it has to map each output neuron to itself.
\end{proof}

\section{Additional Experiments}
\subsection{Details in Datasets, Evaluations and Implementations}
For each dataset, we center-cropped and resized the images, where images in STL-10, LSUN bedroom, and ImageNet were resized to $48\times48$, $64\times64$ and $256\times256$ respectively. 
In ablation study with ResNet architectures we generated 10K images for fast evaluation.In all other cases we used 50K generated samples against real sets for FID calculation.
Lower FID and KID scores and higher PR indicate better performance. 
Adam optimizer was used with learning rate $1e-4$, $\beta_1 = 0$ and $\beta_2 = 0.9$. The length of input noise vector $z$ was set to $128$,  and batch size was fixed to $64$. 
Precision and recall were calculated against test set for CIFAR-10 and validation set for ImageNet.

\subsection{Settings of $\gamma$ in StyleGAN2 Experiments}
\begin{figure}[!ht]
	\centering  
	\includegraphics[width=3.4in,height=2.5in]{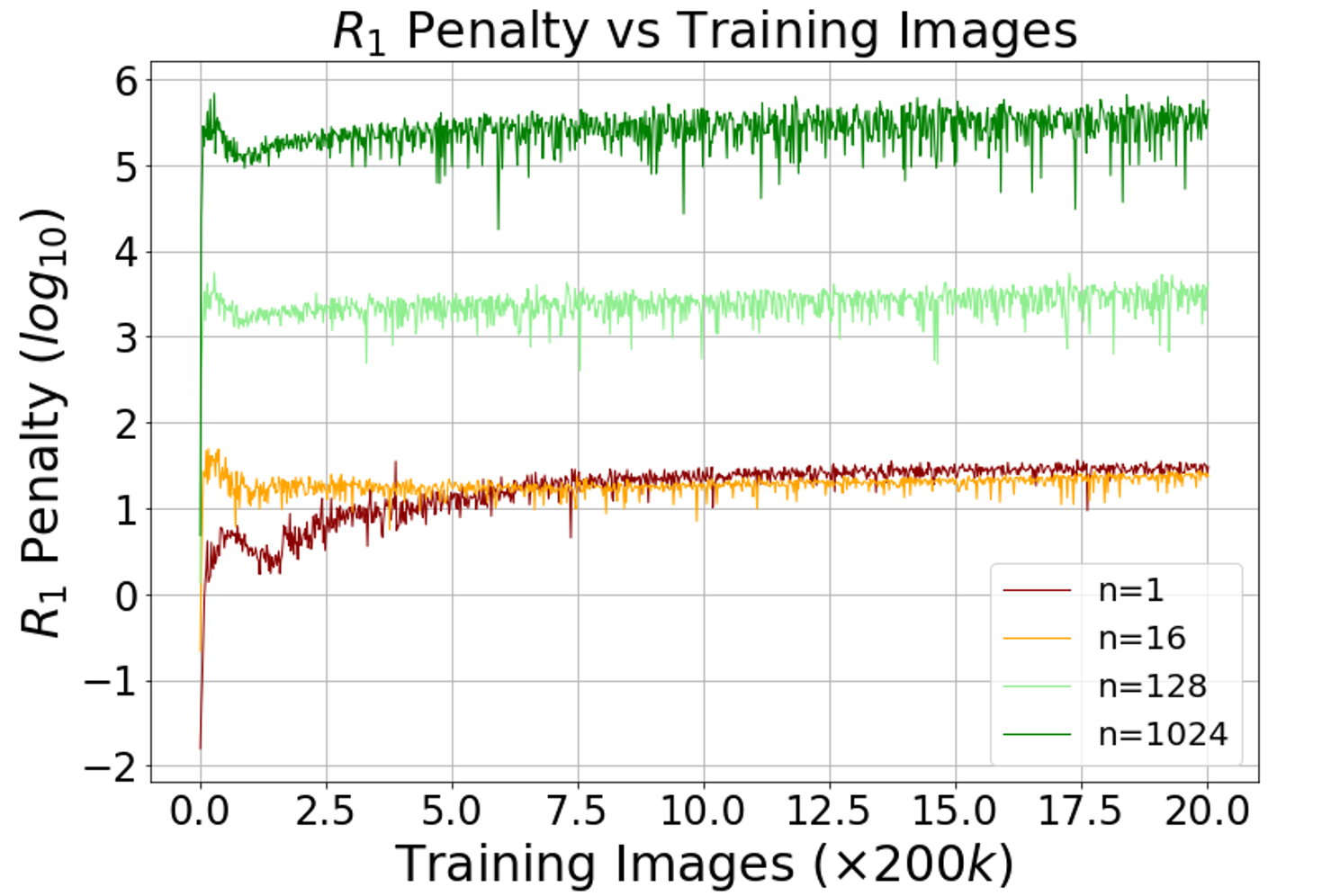}  
	\caption{$R_1$ penalty during training on CIFAR-10 with $n=1,16,128$ and $1024$ using StyleGAN2 architectures.} 
	\label{fig:r1_stylegan2}
\end{figure}

In the experiments we used $\gamma=1e-2,1e-2,1e-4,1e-6$ for $n=1,16,128,1024$ respectively, 
where the total $R_1$ regularization term equals $0.5 \times \gamma \times R_1 $ penalty.  Figure~\ref{fig:r1_stylegan2} shows $R_1$ penalty during training under these settings. 

\subsection{Evaluations of Precision and Recall for Ablations}
\begin{figure}[ht]
	\centering  
	\includegraphics[width=3.3in,height=2.4in]{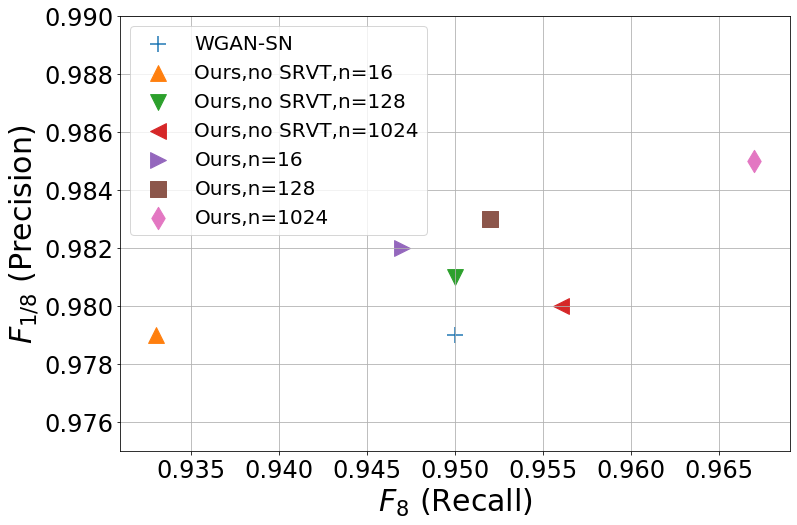}  
	\caption{Precision and recall plot under different settings.} 
	\label{fig:PR}
\end{figure}

In Fig~\ref{fig:PR} we present plots of precision and recall from these settings. 
As we see the setting with the highest dimensional critic output $n=1024$ and with the use of SRVT led to the best results compared to other settings. 
The result also indicates settings with high-dimensional critic output generated more diversified samples.
For WGAN-GP we obtained $(0.850, 0.943)$ recall and precision. 

\subsection{Effect of SRVT in MMD-GAN}
We conducted experiments to validate the effect of SRVT with MMD-GAN objective \cite{li2017mmd}. For implementation we used the authors' default hyper-parameter settings and network architectures.   
\begin{table}[!ht]
	\centering
	\caption{Evaluation of KID(x${10}^3$)$(\downarrow)$ on the effect of SRVT with MMD-GAN objective and DCGAN architectures.}
	\label{tab:MMDGAN}
	\begin{tabular}{l|ccc}
		\hline
		Dimension of critic output $n$  & 16 & 128 & 1024\\
		\hline
		w/o SRVT (Default) & 17(1) & 16(1) & 20(1)\\
		w/ SRVT   & 14(1)  & 13(1) & 16(1)\\
		\hline
	\end{tabular}
\end{table}
From Table~\ref{tab:MMDGAN} one can see SRVT significantly boosts performance for different $n$s. The best result was obtained with $n=128$ (default setup in [28]). We also notice for MMD-GAN, higher $n$ ($1024$) did not improve performance \cite{2018demystifying}, while we have shown our framework can take advantage of higher dimension critic output features.

\end{appendices}

\noindent

		

\end{document}